\documentclass[10pt,twocolumn,letterpaper]{article}

\usepackage[T1]{fontenc}
\usepackage{times} 
\usepackage{graphicx}
\usepackage{booktabs}
\usepackage{amsmath,amssymb}
\usepackage{xcolor}
\usepackage[numbers,sort&compress]{natbib}
\usepackage[hidelinks]{hyperref}
\usepackage{microtype}
\usepackage{enumitem}
\usepackage{subcaption}
\usepackage{multirow}

\setlist[itemize]{leftmargin=*,topsep=2pt,itemsep=1pt,parsep=0pt}

\title{PalmLeaf-VQA: A Multi-Script Visual Question Answering Benchmark for Historical Palm-Leaf Manuscript Understanding Across Diverse Regions}

\author{Nimol Thuon, Jun Du, Panhapin Theang}

\date{University of Science and Technology of China, Université Paris Cité}
\begin{document}
\maketitle

\begin{abstract}
Historical manuscripts remain largely absent from modern vision-language benchmarks, leaving open how well multimodal large language models (MLLMs) handle culturally diverse, degraded, and non-Latin document images. We introduce \textbf{PalmLeaf-VQA}, a multi-script visual question answering benchmark for historical palm-leaf manuscript understanding across South and Southeast Asian traditions. PalmLeaf-VQA contains \textbf{923 curated manuscript images} and \textbf{7,384 question--answer pairs} from eight collection groups: Balinese, Grantha, Jathakam, Kambaramayanam, Kannada, Khmer, Sundanese, and Tamil. Unlike recognition-oriented resources, the benchmark targets manuscript-aware visual reasoning over preservation-relevant cues, including physical condition, line structure, material and coating, binding holes, margins, symbols, drawings, and localized visual artifacts. We evaluate recent proprietary and open-weight MLLMs under open-answer and constrained-answer prompting and provide fine-grained analysis across collections, question categories, and task types. The strongest evaluated model reaches only \textbf{58.00\% exact-match accuracy} on the held-out test split, revealing substantial limitations in current MLLMs for rare-script, degraded-layout, and preservation-oriented document understanding. PalmLeaf-VQA provides a standardized benchmark for advancing culturally grounded and layout-aware multimodal document analysis.
\end{abstract}

\section{Introduction}
\label{sec:intro}
\begin{figure*}[t] \centering \includegraphics[width=\textwidth]{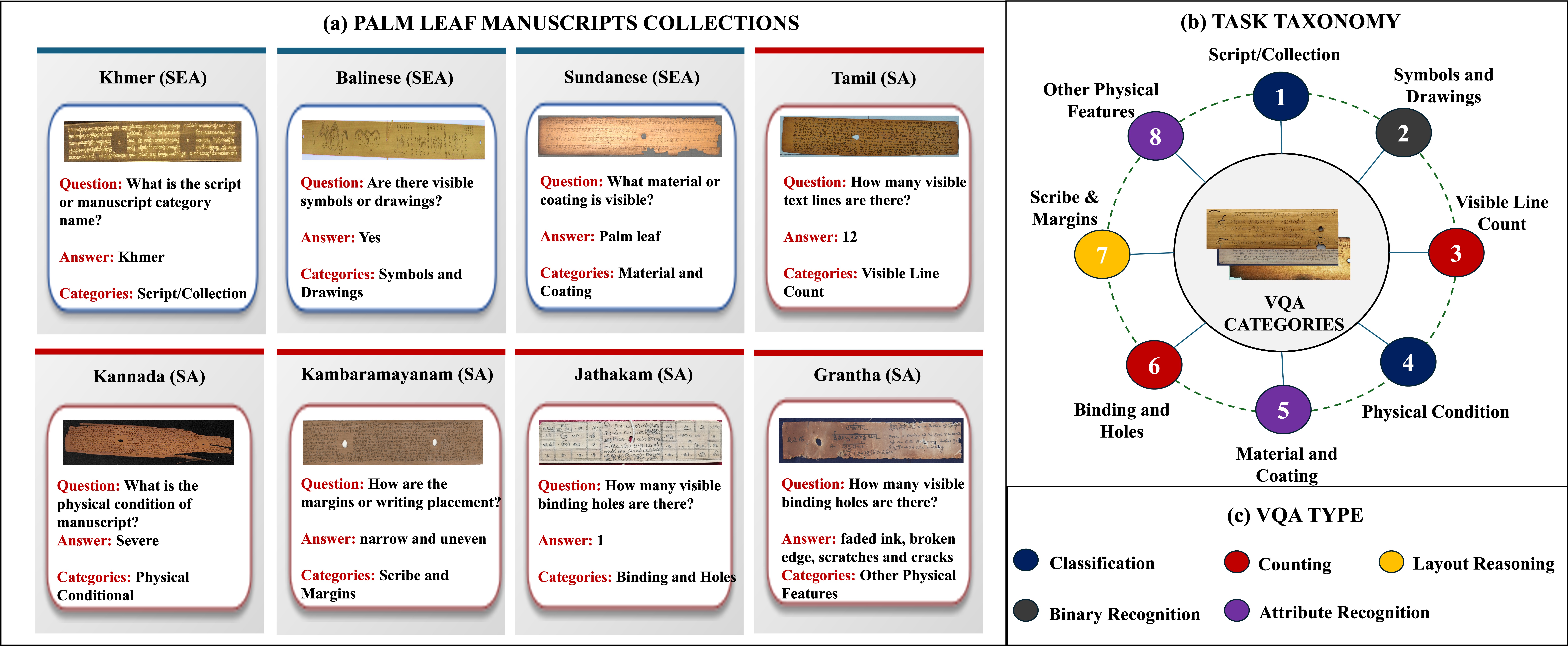} \caption{\textbf{Overview of PalmLeaf-VQA.} (a) Representative palm-leaf manuscript collections from South Asia (SA) and Southeast Asia (SEA), with example visual question-answer pairs. (b) The benchmark covers eight manuscript-aware VQA categories, including script/collection, symbols and drawings, visible line count, physical condition, material and coating, binding and holes, scribe and margins, and other physical features. (c) Questions are grouped into VQA types, including classification, counting, layout reasoning, binary recognition, and attribute recognition.} \label{fig:overview} \end{figure*}
Recent advances in document intelligence and multimodal large language models (MLLMs) have improved the joint modeling of text, images, and layout structure~\cite{xu2020layoutlm,huang2022layoutlmv3,kim2022donut,wang2024docllm}. Historical manuscripts remain a difficult setting for these systems because they must be interpreted not only as written content, but also as material objects with preservation, spatial, and cultural attributes~\cite{nikolaidou2022historicaldatasets,cascianelli2022lam}. Unlike modern printed documents, manuscript images often contain irregular layouts, degraded surfaces, weak contrast, non-standard writing supports, and collection-specific conventions. Existing historical-document benchmarks have advanced layout analysis, binarization, recognition, and writer identification~\cite{simistira2016divahisdb,gruning2018readbad,shen2020hjdataset,cascianelli2022lam}, but they provide limited evaluation of multimodal reasoning beyond transcription and layout parsing.

Palm-leaf manuscripts make this gap especially visible. Used for centuries across South and Southeast Asia, they preserve religious, literary, medical, scientific, and historical knowledge~\cite{wiland2022conservation,chu2023deterioration}. Their long and narrow folios often contain incised or handwritten text arranged around binding holes and constrained margins, while cracks, stains, fading, coatings, biological damage, symbols, and drawings further complicate visual interpretation~\cite{kesiman2016amadi,valy2017sleukrith,suryani2017sundanese,kesiman2018benchmarking}. Many collections also use rare or non-Latin scripts with limited digital resources and weak representation in general-purpose pretraining. Although recent work has improved enhancement, segmentation, layout analysis, recognition, and transcription for specific palm-leaf traditions~\cite{nair2023hmplmd,jailingeswari2024thplmd,jyothi2021grantha,sajjan2025kannada,thuon2022improving,thuon2024khmerformer}, broader MLLM reasoning over manuscript structure, material condition, and preservation cues remains underexplored.

This limitation is not addressed by current VQA and MLLM benchmarks. Existing resources evaluate natural images, scene text, scanned documents, infographics, charts, and PDF-style content~\cite{antol2015vqa,singh2019textvqa,mathew2021docvqa,mathew2022infographicvqa,masry2022chartqa}, while more recent evaluations probe OCR, multimodal reasoning, and document understanding at larger scale~\cite{fu2023mme,liu2024mmbench,li2024seedbench,yue2024mmmu,liu2024ocrbench}. However, these benchmarks rarely require models to reason jointly over rare scripts, physical materials, irregular layouts, and degradation. As a result, they provide only limited evidence about whether current MLLMs can understand historical manuscripts rather than modern document types that are heavily represented in web-scale training data.

To address this gap, we introduce \textbf{PalmLeaf-VQA}, a multi-script visual question answering benchmark for historical palm-leaf manuscript understanding across South and Southeast Asia. The benchmark contains \textbf{923 curated images} and \textbf{7,384 newly annotated question--answer pairs} across eight collection groups: Balinese, Grantha, Jathakam, Kambaramayanam, Kannada, Khmer, Sundanese, and Tamil. PalmLeaf-VQA uses image-disjoint training, validation, and test splits to prevent visual overlap across subsets. It covers eight manuscript-aware categories, including script identification, physical condition, line counting, symbols and drawings, material and coating, binding and holes, scribe and margins, and other physical features. These categories extend evaluation beyond OCR and transcription toward classification, counting, layout reasoning, binary recognition, and preservation-aware attribute understanding.

We benchmark recent proprietary and open-weight MLLMs without task-specific fine-tuning under open-answer and constrained-answer settings. The findings show that historical palm-leaf manuscript understanding remains far from solved for current general-purpose MLLMs.

Our contributions are threefold:
\begin{itemize}
\item We introduce \textbf{PalmLeaf-VQA}, a multi-script VQA benchmark that evaluates historical palm-leaf manuscripts as textual documents and physical artifacts.
\item We provide \textbf{923 curated images} and \textbf{7,384 newly annotated QA pairs} across eight collection groups, with image-disjoint training, validation, and test splits.
\item We establish a systematic zero-shot evaluation of recent proprietary and open-weight MLLMs across eight manuscript-aware categories, revealing persistent failures in rare-script recognition, degradation-aware reasoning, and preservation-oriented visual understanding.
\end{itemize}

\section{Related Work}
\label{sec:related_work}
\begin{table}[t]
\centering
\setlength{\tabcolsep}{1.8pt}
\scriptsize
\begin{tabular}{@{}llccccc@{}}
\toprule
\textbf{Benchmark} & \textbf{Coverage} & \textbf{Hist.} & \textbf{Low-R} & \textbf{Layout} & \textbf{VQA} & \textbf{Pres.} \\
\midrule
VQA~\cite{antol2015vqa} & Natural images & -- & -- & -- & \(\checkmark\) & -- \\
TextVQA~\cite{singh2019textvqa} & Scene text & -- & -- & \(\circ\) & \(\checkmark\) & -- \\
DocVQA~\cite{mathew2021docvqa} & Scanned docs & -- & -- & \(\checkmark\) & \(\checkmark\) & -- \\
InfographicVQA~\cite{mathew2022infographicvqa} & Infographics & -- & -- & \(\checkmark\) & \(\checkmark\) & -- \\
ChartQA~\cite{masry2022chartqa} & Charts & -- & -- & \(\checkmark\) & \(\checkmark\) & -- \\
DUDE~\cite{tito2023dude} & Multi-domain docs & -- & -- & \(\checkmark\) & \(\checkmark\) & -- \\
MMLongBench-Doc~\cite{ma2024mmlongbenchdoc} & Long docs & -- & -- & \(\checkmark\) & \(\checkmark\) & -- \\
SEA-Vision~\cite{yue2026seavision} & SEA docs/scene text & -- & \(\checkmark\) & \(\checkmark\) & \(\checkmark\) & -- \\
OCRBench~\cite{liu2024ocrbench} & OCR tasks & -- & -- & \(\checkmark\) & \(\circ\) & -- \\
MMDocBench~\cite{zhu2024mmdocbench} & Fine-grained docs & -- & -- & \(\checkmark\) & \(\circ\) & -- \\
ICFHR (non-QA)~\cite{kesiman2018icfhr} & Palm-leaf, 3 groups & \(\checkmark\) & \(\checkmark\) & \(\checkmark\) & -- & \(\circ\) \\
\textbf{PalmLeaf-VQA} & \textbf{Palm-leaf, 8 groups} & \(\checkmark\) & \(\checkmark\) & \(\checkmark\) & \(\checkmark\) & \(\checkmark\) \\
\bottomrule
\end{tabular}
\caption{\textbf{Compact comparison of benchmark scope.}
The table does not rank dataset quality; it highlights whether each benchmark explicitly targets historical materials (Hist.), rare or low-resource scripts/languages (Low-R), layout reasoning, visual question answering (VQA), and material or preservation cues (Pres.). \(\checkmark\): primary focus; \(\circ\): partial or indirect coverage.}
\label{tab:benchmark_comparison}
\end{table}

\subsection{Document VQA and MLLM Evaluation}

Visual question answering has expanded from natural-image reasoning~\cite{antol2015vqa} to text-rich and document-centric settings. TextVQA evaluates reading and reasoning over scene text~\cite{singh2019textvqa}, while DocVQA introduced question answering over scanned document images~\cite{mathew2021docvqa}. Subsequent benchmarks broadened evaluation to infographics, charts, multi-domain documents, and long PDF-style inputs~\cite{mathew2022infographicvqa,masry2022chartqa,tito2023dude,ma2024mmlongbenchdoc}. Recent benchmarks such as OCRBench, OCRBench v2, and MMDocBench further probe OCR, localization, fine-grained perception, and reasoning over complex document layouts~\cite{liu2024ocrbench,zhu2024mmdocbench,fu2024ocrbenchv2improvedbenchmark}.

More general MLLM evaluations, including MME, MMBench, SEED-Bench, and MMMU, assess multimodal perception, reasoning, knowledge, and instruction following across broad visual domains~\cite{fu2023mme,liu2024mmbench,li2024seedbench,yue2024mmmu}. Multilingual and culturally broader benchmarks have also begun to extend evaluation beyond high-resource languages~\cite{liu2021marvl,romero2024cvqa,yue2026seavision}. However, these resources primarily emphasize natural scenes, modern documents, charts, or web-style content. They rarely evaluate historical manuscripts whose interpretation depends jointly on rare scripts, physical materials, irregular layouts, degradation, and preservation-related evidence.

\subsection{Historical and Palm-Leaf Manuscript Analysis}

Historical document analysis has long studied binarization, layout parsing, baseline detection, handwriting recognition, and writer identification for degraded archival materials~\cite{simistira2016divahisdb,gruning2018readbad,nikolaidou2022historicaldatasets,cascianelli2022lam}. Palm-leaf manuscripts form a particularly challenging subset because of their elongated geometry, non-paper writing supports, low-contrast strokes, binding holes, coatings, surface damage, and visually complex non-Latin scripts.

Prior work has introduced resources and methods for Balinese, Khmer, Sundanese, Malayalam, Tamil, Grantha, and Kannada palm-leaf manuscripts~\cite{kesiman2016amadi,valy2017sleukrith,suryani2017sundanese,nair2023hmplmd,jailingeswari2024thplmd,jyothi2021grantha,sajjan2025kannada}. The ICFHR palm-leaf benchmarks evaluated binarization, text-line segmentation, glyph recognition, word recognition, and transliteration across several Southeast Asian collections~\cite{kesiman2018icfhr,kesiman2018benchmarking}. More recent studies have addressed enhancement, dataset expansion, recognition, layout analysis, damage segmentation, and text-line segmentation~\cite{thuon2022improving,thuon2024khmerformer,thuon2024generate,thuon2025dual_loop,thuon2025palmlay,wang2024damage,sivan2026leafocrline}. These resources, however, primarily treat manuscripts as recognition, segmentation, or restoration targets. PalmLeaf-VQA instead evaluates whether MLLMs can answer broader visual questions about manuscripts as both textual documents and physical artifacts.

\section{PalmLeaf-VQA Benchmark}
\label{sec:benchmark}
\begin{table}[t] 
\centering 
\setlength{\tabcolsep}{3pt} 
\scriptsize 
\begin{tabular}
{@{}lrrrrr@{}} \toprule \textbf{Collection} & \textbf{Train} & \textbf{Val} & \textbf{Test} & \textbf{Images} & \textbf{QA} \\ \midrule Balinese & 60 & 10 & 30 & 100 & 800 \\ Grantha & 42 & 7 & 21 & 70 & 560 \\ Jathakam & 59 & 10 & 30 & 99 & 792 \\ Kambaramayanam & 48 & 8 & 24 & 80 & 640 \\ Kannada & 154 & 26 & 77 & 257 & 2,056 \\ Khmer & 93 & 15 & 47 & 155 & 1,240 \\ Sundanese & 37 & 6 & 18 & 61 & 488 \\ Tamil & 60 & 10 & 31 & 101 & 808 \\ \midrule \textbf{Total} & \textbf{553} & \textbf{92} & \textbf{278} & \textbf{923} & \textbf{7,384} \\ 
\bottomrule 
\end{tabular} 
\caption{\textbf{PalmLeaf-VQA dataset statistics by collection group.} Splits are image-disjoint. Each image has eight question-answer pairs.} 
\label{tab:dataset_stats} 
\end{table}
PalmLeaf-VQA reformulates curated palm-leaf manuscript collections as a unified benchmark for manuscript-aware visual question answering. Figure~\ref{fig:overview} summarizes its collections, question categories, and task types, while Table~\ref{tab:dataset_stats} reports collection-level statistics.

\subsection{Source Collections and Curation}

PalmLeaf-VQA draws on public and research-oriented manuscript resources from South and Southeast Asia, including AMADI\_LontarSet~\cite{kesiman2016amadi}, the SleukRith Set~\cite{valy2017sleukrith}, Sundanese palm-leaf manuscripts~\cite{suryani2017sundanese}, and the ICFHR 2018 resources~\cite{kesiman2018icfhr}. Additional sources cover Grantha, Jathakam, Kambaramayanam, Kannada, Malayalam, and Tamil manuscript traditions ~\cite{jyothi2021grantha,sajjan2025kannada,nair2023hmplmd,jailingeswari2024thplmd}.

Because the original datasets were developed for recognition, segmentation, layout analysis, and transcription, they were not merged directly. We removed unusable captures, unrelated content, severe framing errors, and duplicate or near-duplicate images. Cropping and framing normalization were applied only when necessary, while preserving manuscript content, degradation, binding structures, and material characteristics.

The final benchmark contains eight collection groups: Balinese, Grantha, Jathakam, Kambaramayanam, Kannada, Khmer, Sundanese, and Tamil. We use \emph{collection group} because these labels refer to source collections or manuscript traditions and do not always correspond one-to-one with distinct scripts.

\subsection{Question Taxonomy and Answer Design}
\label{sec:question_taxonomy}

Each image is paired with eight questions covering script or collection identity, physical condition, visible line count, symbols and drawings, material and coating, binding and holes, scribe and margins, and other physical features. With one question from each category per image, the taxonomy is balanced and decomposes aggregate performance into fine-grained document and visual capabilities~\cite{fu2024ocrbenchv2improvedbenchmark,li2024naturalbench,zhu2024mmdocbench}.

\begin{figure}[!t]
\centering
\includegraphics[width=\columnwidth]{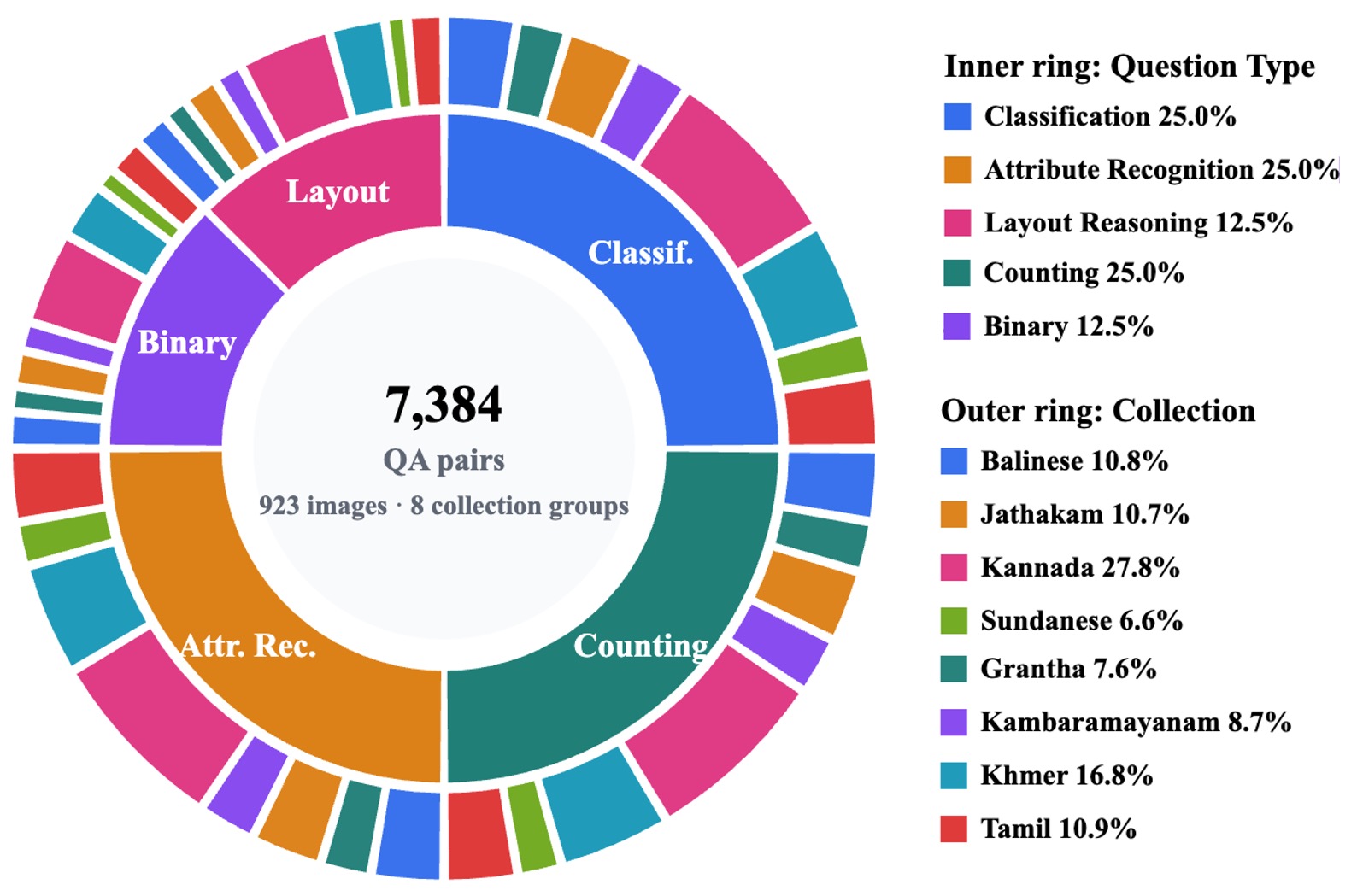}
\caption{\textbf{Task-type and collection distribution in PalmLeaf-VQA.}
The nested donut summarizes 7,384 question--answer pairs from 923 manuscript images. The inner ring aggregates the eight equally represented question categories into five VQA task types, while the outer ring decomposes each task-type segment across the eight collection groups.}
\label{fig:task_collection_distribution}
\end{figure}

Questions are grounded in visually observable evidence rather than external historical knowledge or full-text translation. The eight categories are mapped to five VQA task types: classification, counting, binary recognition, layout reasoning, and attribute recognition. Figure~\ref{fig:task_collection_distribution} complements the collection statistics in Table~\ref{tab:dataset_stats} by visualizing this two-level benchmark composition. The inner ring aggregates the eight equally represented question categories into the five task types, while the outer ring shows how the question--answer pairs are distributed across the eight collection groups. The larger Kannada and Khmer segments reflect their larger image counts, whereas every image still contributes the same eight question categories.

Answers may be single-label or multi-label. Categorical properties use canonical vocabularies, binary properties use standardized labels, and visible-line counts are represented numerically. Multi-label answers are stored as unordered canonical sets. The same reference annotations are used for open-answer and constrained-answer evaluation.

\subsection{Annotation and Validation}

The annotation team comprised 12 annotators from South, Southeast, and East Asia with experience in palm-leaf manuscripts, historical document analysis, or the represented manuscript traditions. Assignments were matched to regional, script, or collection familiarity whenever possible, consistent with culturally grounded VQA practices that emphasize relevant linguistic and cultural expertise~\cite{romero2024cvqa}. 
Each question--answer pair was written by one annotator and independently reviewed by a second team member. Disagreements were resolved through discussion and adjudication by an experienced reviewer, while unresolved cases were excluded.

Annotations were checked for visual answerability, category consistency, and agreement with the image evidence. Script names, categorical attributes, binary labels, numeric counts, punctuation, and multi-label formatting were standardized. Full guidelines and agreement statistics are provided in the supplementary material.

\subsection{Quality Control, Splits, and Release}

Images were inspected for visibility, framing, relevance, and capture quality. Exact duplicates were detected through file-level matching, while near-duplicates were screened using perceptual similarity, source metadata, and manual inspection.

The benchmark contains \textbf{923 images} and \textbf{7,384 question--answer pairs}, divided into image-disjoint training, validation, and test splits of 553, 92, and 278 images. Each image contributes one question per category, making the benchmark balanced by question category.

Where reliable identifiers were available, images from the same physical manuscript or capture sequence were assigned to the same split. Otherwise, source folders, filenames, metadata, and near-duplicate clusters were used as grouping evidence. We therefore describe the benchmark as image-disjoint rather than fully manuscript-disjoint. For restricted sources, we will release annotations, source identifiers, retrieval instructions, and processing scripts.
\section{Evaluation Protocol}
\label{sec:evaluation_protocol}

\subsection{Task and Test Setting}

Following standard VQA formulations~\cite{antol2015vqa,singh2019textvqa,mathew2021docvqa} and recent document-focused multimodal benchmarks~\cite{ma2024mmlongbenchdoc,zhu2024mmdocbench,fu2024ocrbenchv2improvedbenchmark}, a model receives a manuscript image $I_i$ and question $q_i$ and predicts an answer $\hat{a}_i$. Evaluation is conducted on \textbf{2,224 question--answer pairs} from \textbf{278 image-disjoint test images}.

Models receive only the image and question, without OCR transcripts, source metadata, ground-truth labels, preservation annotations, or demonstrations. The validation split is used to finalize normalization rules and prompt templates, which are fixed before test evaluation. The training split is not used in the reported zero-shot experiments.

All models use the same test examples and scoring pipeline. Model identifiers, API access dates, image-resolution settings, decoding parameters, and implementation details are provided in the supplementary material.

\subsection{Prompting Conditions}

We evaluate free-form and constrained generation, two complementary settings commonly used in recent multimodal benchmarks~\cite{liu2024mmbench,yue2024mmmu,yue2024mmmupro,ma2024mmlongbenchdoc,fu2024ocrbenchv2improvedbenchmark}. In the \textbf{open-answer} setting, models generate a free-form response without category-specific answer choices. In the \textbf{constrained-answer} setting, models additionally receive the required output format and, when applicable, the canonical answer vocabulary.

The image, question, and reference annotation remain unchanged across conditions. This comparison separates visual-understanding errors from lexical and formatting variation. We use \emph{closed} only as a compact label in figures and tables. All models are evaluated without task-specific fine-tuning.

\subsection{Answer Normalization and Metrics}

Predictions and references are normalized through lowercasing, whitespace and punctuation cleanup, canonicalization of script and collection names, standardization of binary labels, and numeric parsing. Multi-label answers are represented as unordered canonical sets, and invalid or unparseable outputs are counted as incorrect.

Our primary metric is \textbf{exact-match accuracy}:
\begin{equation}
\mathrm{EM}
=
\frac{1}{N}
\sum_{i=1}^{N}
\mathbf{1}
\left[
\mathrm{norm}(\hat{a}_i)=\mathrm{norm}(a_i)
\right],
\label{eq:accuracy}
\end{equation}
where $N$ is the number of test examples, $\hat{a}_i$ is the predicted answer, $a_i$ is the reference answer, and $\mathrm{norm}(\cdot)$ denotes answer normalization.

For multi-label questions, exact match requires equality between the predicted and reference sets. We additionally report precision, recall, F1, and Jaccard similarity to capture partial correctness. For line-count questions, we report exact accuracy and \emph{within-one accuracy}. Complete parsing rules and metric definitions are provided in the supplementary material.

\subsection{Fine-Grained and Human Evaluation}

Following recent diagnostic benchmarks that report results across skills, document types, and evidence granularity~\cite{zhu2024mmdocbench,li2024naturalbench,ouyang2025omnidocbench}, we analyze performance by collection group, question category, task type, answer cardinality, and prompting condition. This includes separate results for single-answer and multi-answer questions and for classification, counting, binary recognition, layout reasoning, and attribute recognition.

Human performance is evaluated on a balanced test subset under the same constrained-answer protocol used for the models. Three independent responses are collected per question. Human EM and F1 are averaged over all 768 responses, while W1 is computed over the 96 responses to the 32 visible-line-count questions. Response consistency is measured using mean pairwise exact agreement after normalization. Full details are provided in the supplementary material.

\section{Experiments}
\label{sec:experiments}

We evaluate recent proprietary and open-weight models on the PalmLeaf-VQA held-out test split using the protocol defined in Section~\ref{sec:evaluation_protocol}. Following established MLLM and document-understanding benchmarks~\cite{liu2024mmbench,liu2024ocrbench,ma2024mmlongbenchdoc}, we use exact-match accuracy (EM) as the primary metric and report mean answer F1 and within-one line-count accuracy (W1) as complementary measures. All quantitative results are obtained from directly measured prediction runs without task-specific fine-tuning.

\subsection{Models and Experimental Setup}

Our evaluation includes proprietary MLLMs, open-weight vision-language models, and text-only controls. The proprietary systems are Gemini 3.5 Flash and Gemini 3.1 Flash Lite~\cite{google2026gemini35,google2026gemini31}, and ChatGPT 5.5~\cite{openai2026gpt55}. The open-weight multimodal models include Qwen3-VL-8B~\cite{Qwen3-VL}, Qwen2.5-VL-7B and Qwen2.5-VL-3B~\cite{Qwen2.5-VL}, InternVL3.5-8B and InternVL3.5-4B~\cite{wang2025internvl3}, and Gemma-3-4B-IT~\cite{gemma3technicalreport}. We additionally include Llama-3.1-8B and Llama-3.2-3B as text-only controls~\cite{dubey2024llama}; these models receive the question but not the manuscript image.

Every model is evaluated on the same 2,224 test questions under the open-answer and constrained-answer conditions described in Section~\ref{sec:evaluation_protocol}. We use the same image inputs, normalization procedure, and scoring implementation across systems. Exact model identifiers, API access dates, decoding parameters, image-resolution settings, prompt templates, hardware configurations, and inference details are provided in the supplementary material.
\begin{table}[t]
\centering
\setlength{\tabcolsep}{2.8pt}
\scriptsize
\begin{tabular}{@{}llrrr@{\hspace{0.8em}}rrr@{}}
\toprule
 &  & \multicolumn{3}{c}{\textbf{Constrained-answer}} & \multicolumn{3}{c}{\textbf{Free-form/open}} \\
\cmidrule(lr){3-5} \cmidrule(lr){6-8}
\textbf{Type} & \textbf{Model} & \textbf{EM} & \textbf{F1} & \textbf{W1} & \textbf{EM} & \textbf{F1} & \textbf{W1} \\
\midrule
\multirow{8}{*}{Open-weight}
& Qwen3-VL-8B & 44.02 & 54.73 & \textbf{71.22} & 21.49 & 21.80 & 68.60 \\
& Qwen2.5-VL-7B & 31.03 & 37.41 & 61.87 & 13.62 & 13.95 & \textbf{71.92} \\
& Qwen2.5-VL-3B & 35.21 & 41.71 & 56.29 & 15.15 & 15.61 & 54.33 \\
& InternVL3.5-8B & 44.96 & 54.48 & 62.05 & 24.78 & 25.02 & 61.51 \\
& InternVL3.5-4B & \textbf{46.00} & \textbf{54.93} & 59.71 & \textbf{24.96} & \textbf{25.12} & 65.65 \\
& Llama-3.1-8B & 31.65 & 34.05 & 34.35 & 8.00 & 8.00 & 3.96 \\
& Llama-3.2-3B & 26.26 & 26.26 & 23.92 & 7.91 & 7.91 & 0.00 \\
& Gemma-3-4B-IT & 30.94 & 40.13 & 14.75 & 19.29 & 21.58 & 29.50 \\
\midrule
\multirow{3}{*}{Proprietary}
& Gemini 3.5 Flash & \textbf{58.00} & \textbf{67.25} & \textbf{93.88} & \textbf{43.03} & \textbf{43.31} & \textbf{95.13} \\
& ChatGPT 5.5 & 55.40 & 66.17 & 93.53 & 40.50 & 39.16 & 94.68 \\
& Gemini 3.1 FL & 47.53 & 58.80 & 74.28 & 39.52 & 39.58 & 75.72 \\
\bottomrule
\end{tabular}
\caption{\textbf{Full model results on PalmLeaf-VQA under constrained-answer and free-form/open prompting.} Values are percentages on the held-out test split. Type distinguishes open-weight models from proprietary systems. EM denotes exact-match accuracy, F1 denotes mean answer F1, and W1 denotes within-one accuracy for line-count questions. Bold marks the best result within each access type. FL denotes Flash-Lite.}
\label{tab:main_results_full}
\end{table}

\begin{table}[t]
\centering
\setlength{\tabcolsep}{3pt}
\scriptsize
\begin{tabular}{@{}lrrrrrrrrr@{}}
\toprule
\textbf{Model} & \textbf{Bal} & \textbf{Gra} & \textbf{Jat} & \textbf{Kam} & \textbf{Kan} & \textbf{Khm} & \textbf{Sun} & \textbf{Tam} & \textbf{Avg.} \\
\midrule

Qwen3-VL-8B & 47.9 & 38.7 & 38.3 & 47.9 & 41.6 & 42.6 & 43.8 & 54.8 & 44.0 \\
Qwen2.5-VL-7B & 30.0 & 29.2 & 46.2 & 28.6 & 25.0 & 32.2 & 29.2 & 34.7 & 31.0 \\
Qwen2.5-VL-3B & 27.5 & 41.1 & 33.3 & 38.0 & 34.9 & 43.4 & 27.8 & 31.0 & 35.2 \\
InternVL3.5-8B & 50.8 & \textbf{67.3} & 37.9 & 48.4 & 38.1 & 46.8 & 43.8 & 43.1 & 45.0 \\
InternVL3.5-4B & 49.2 & 50.0 & 42.9 & 49.0 & 47.7 & 37.0 & 39.6 & 54.0 & 46.0 \\
Llama-3.1-8B & 34.2 & 40.5 & 36.2 & 32.8 & 18.7 & 37.5 & 24.3 & 45.6 & 31.6 \\
Llama-3.2-3B & 25.0 & 28.0 & 33.8 & 24.0 & 13.8 & 35.6 & 18.8 & 41.9 & 26.3 \\
Gemma-3-4B-IT & 31.2 & 18.5 & 21.7 & 28.6 & 41.2 & 28.2 & 31.2 & 28.2 & 30.9 \\
\midrule

Gemini 3.5 Flash & \textbf{65.8} & 64.3 & \textbf{58.3} & \textbf{51.0} & 57.0 & \textbf{56.9} & \textbf{61.1} & 53.6 & 58.0 \\
ChatGPT 5.5 & 62.1 & 48.2 & 54.2 & 47.4 & \textbf{59.4} & 51.3 & 59.0 & \textbf{55.2} & 55.4 \\
Gemini 3.1 FL & 57.1 & 42.9 & 41.7 & 38.5 & 48.7 & 47.6 & 52.8 & 48.0 & 47.5 \\
\bottomrule
\end{tabular}
\caption{\textbf{Collection-level exact-match accuracy under constrained-answer prompting.} Values are percentages on the held-out test split. Avg. matches constrained EM in Table~\ref{tab:main_results_full}. Column abbreviations: Balinese (Bal), Grantha (Gra), Jathakam (Jat), Kambaramayanam (Kam), Kannada (Kan), Khmer (Khm), Sundanese (Sun), and Tamil (Tam).}
\label{tab:script_results_constrained}
\end{table}

\begin{table}[t]
\centering
\setlength{\tabcolsep}{4.5pt}
\scriptsize
\begin{tabular}{@{}lrrrr@{}}
\toprule
\textbf{Evaluator} & \textbf{Questions} & \textbf{EM} & \textbf{F1} & \textbf{W1} \\
\midrule
Human participants & 256 & \textbf{87.50} & \textbf{92.40} & \textbf{96.88} \\
Gemini 3.5 Flash & 256 & 58.59 & 67.07 & 93.75 \\
ChatGPT 5.5 & 256 & 59.77 & 69.89 & 90.63 \\
\bottomrule
\end{tabular}
\caption{\textbf{Human and model performance on a balanced 256-question subset.}
Human EM and F1 are averaged over 768 responses, while human W1 is computed over 96 responses to the 32 line-count questions. Model EM and F1 are computed over 256 questions, while model W1 is computed over the corresponding 32 line-count questions. All evaluators use the constrained-answer setting. Values are percentages.}
\label{tab:human_baseline}
\end{table}
\subsection{Overall Performance}

Table~\ref{tab:main_results_full} yields three main observations. First, proprietary multimodal models lead the benchmark. Gemini 3.5 Flash achieves the strongest constrained-answer result with \textbf{58.00\% EM}, \textbf{67.25\% F1}, and \textbf{93.88\% W1}, followed by ChatGPT 5.5 at \textbf{55.40\% EM} and Gemini 3.1 Flash Lite at \textbf{47.53\% EM}.

Second, among open-weight vision-language models, InternVL3.5-4B performs best under constrained-answer prompting with \textbf{46.00\% EM}, slightly ahead of InternVL3.5-8B (\textbf{44.96\%}) and Qwen3-VL-8B (\textbf{44.02\%}). This result indicates that parameter count alone is not decisive for palm-leaf manuscript understanding. Third, the text-only Llama controls perform substantially worse than image-conditioned models, confirming that the manuscript image provides essential evidence beyond the question text.

The same ranking largely holds under open-answer prompting. Gemini 3.5 Flash again performs best with \textbf{43.03\% EM}, followed by ChatGPT 5.5 (\textbf{40.50\%}) and Gemini 3.1 Flash Lite (\textbf{39.52\%}). Even so, the best constrained-answer model reaches only \textbf{58.00\% EM}, leaving substantial headroom for future progress and showing that PalmLeaf-VQA is far from saturated by current general-purpose MLLMs.

The gap between EM and F1 further suggests that models often recover partially correct information without producing the full canonical answer. For Gemini 3.5 Flash, this gap is more than nine points under constrained prompting, which is especially relevant for preservation-oriented questions where a response may capture one valid attribute while omitting another required one.

\begin{figure*}
\centering
\begin{subfigure}{0.49\linewidth}
\centering
\includegraphics[width=\linewidth]{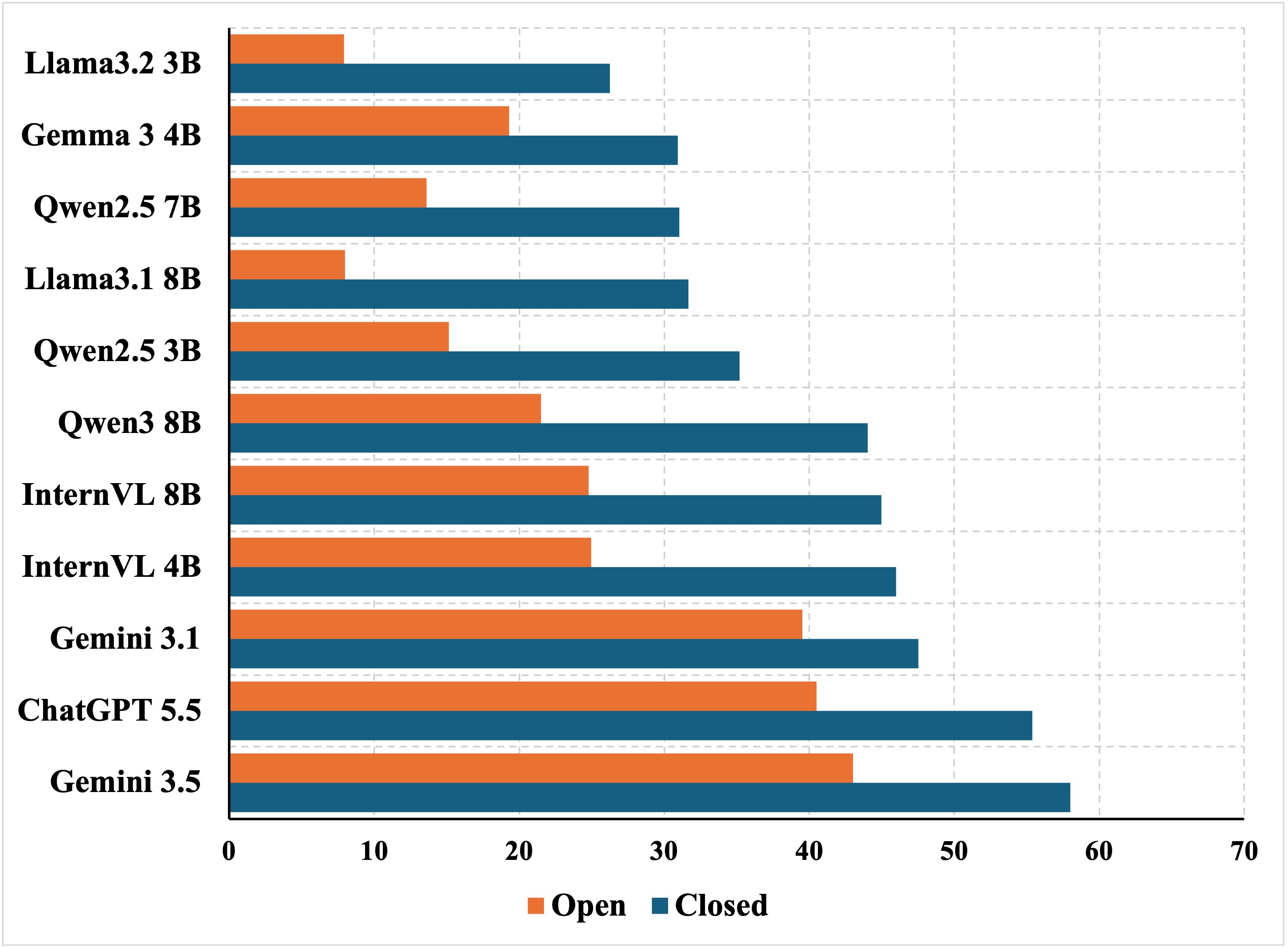}
\caption{Open- versus constrained-answer exact-match accuracy (\%).}
\label{fig3:a}
\end{subfigure}
\hfill
\begin{subfigure}{0.49\linewidth}
\centering
\includegraphics[width=\linewidth]{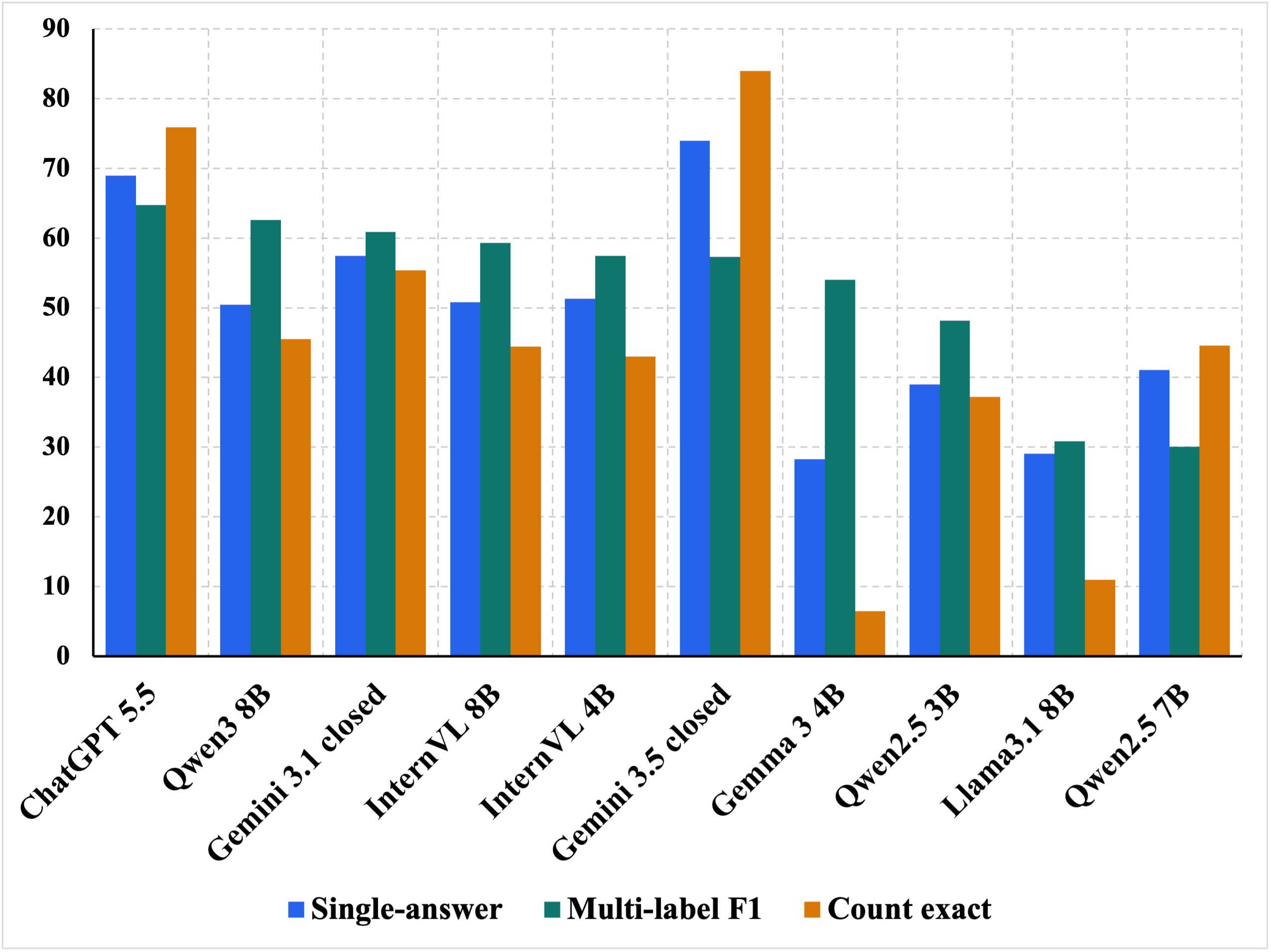}
\caption{Performance by answer type (\%).}
\label{fig3:b}
\end{subfigure}
\caption{\textbf{Performance under different prompting and answer settings.}
(a) Exact-match accuracy under open-answer and constrained-answer prompting.
(b) Performance by answer type, including single-answer EM, multi-answer F1, and exact line-count accuracy.}
\label{fig:prompt_answer_analysis}
\end{figure*}

\subsection{Effect of Prompting and Answer Cardinality}

Figure~\ref{fig:prompt_answer_analysis} shows two consistent trends. First, constrained-answer prompting improves performance for every evaluated model. Gemini 3.5 Flash improves from \textbf{43.03\%} to \textbf{58.00\% EM}, a gain of \textbf{14.97} percentage points. ChatGPT 5.5 improves by \textbf{14.90} points, while Gemini 3.1 Flash Lite gains \textbf{8.01} points. Larger gains are observed for several open-weight models, including Qwen3-VL-8B (\textbf{22.53} points), InternVL3.5-8B (\textbf{20.18} points), and InternVL3.5-4B (\textbf{21.04} points). Across all eleven paired comparisons, constrained prompting improves EM by an average of \textbf{17.52} percentage points.

This improvement shows that part of the challenge lies in answer-space variance: under open-answer prompting, models often produce explanations, paraphrases, or near-miss descriptions that do not map cleanly to the canonical label space. However, output control is only part of the story. Even with constrained prompting, the strongest model remains incorrect on more than two-fifths of the test questions, indicating that the main bottleneck is still visual understanding rather than answer formatting alone.

Second, multi-answer questions are consistently harder than single-answer questions, as shown in Figure~\ref{fig:prompt_answer_analysis}(b). These cases require models to recover the complete set of valid attributes rather than only the most salient one. Typical errors include missing secondary attributes and adding plausible but unsupported labels, which lowers exact-match accuracy even when part of the prediction is correct.

\subsection{Collection-Level Performance}

Table~\ref{tab:script_results_constrained} reveals substantial variation across collection groups. Gemini 3.5 Flash achieves the strongest average result and leads on five of the eight collections. Its highest score is obtained on Balinese (\textbf{65.8\%}), followed by Grantha (\textbf{64.3\%}) and Sundanese (\textbf{61.1\%}), while Kambaramayanam is the most difficult group at \textbf{51.0\%}. The resulting spread of \textbf{14.8} points indicates that aggregate performance masks meaningful collection-specific differences.

The strongest model is not always the same across collections. InternVL3.5-8B obtains the best Grantha result at \textbf{67.3\%}, despite a lower overall average, while ChatGPT 5.5 performs best on Kannada and Tamil with \textbf{59.4\%} and \textbf{55.2\%}, respectively. This pattern suggests that collection-level performance depends on properties such as script appearance, line density, page geometry, acquisition quality, and preservation condition, rather than on overall model scale alone.

Collection difficulty is also not explained simply by sample size. Kannada is the largest group but is not the easiest, whereas the smaller Balinese and Grantha groups are among the strongest for several models. These results motivate reporting collection-level accuracy in addition to the global leaderboard.

\subsection{Category-Level Performance}

\begin{figure}[t]
\centering
\includegraphics[width=\linewidth]{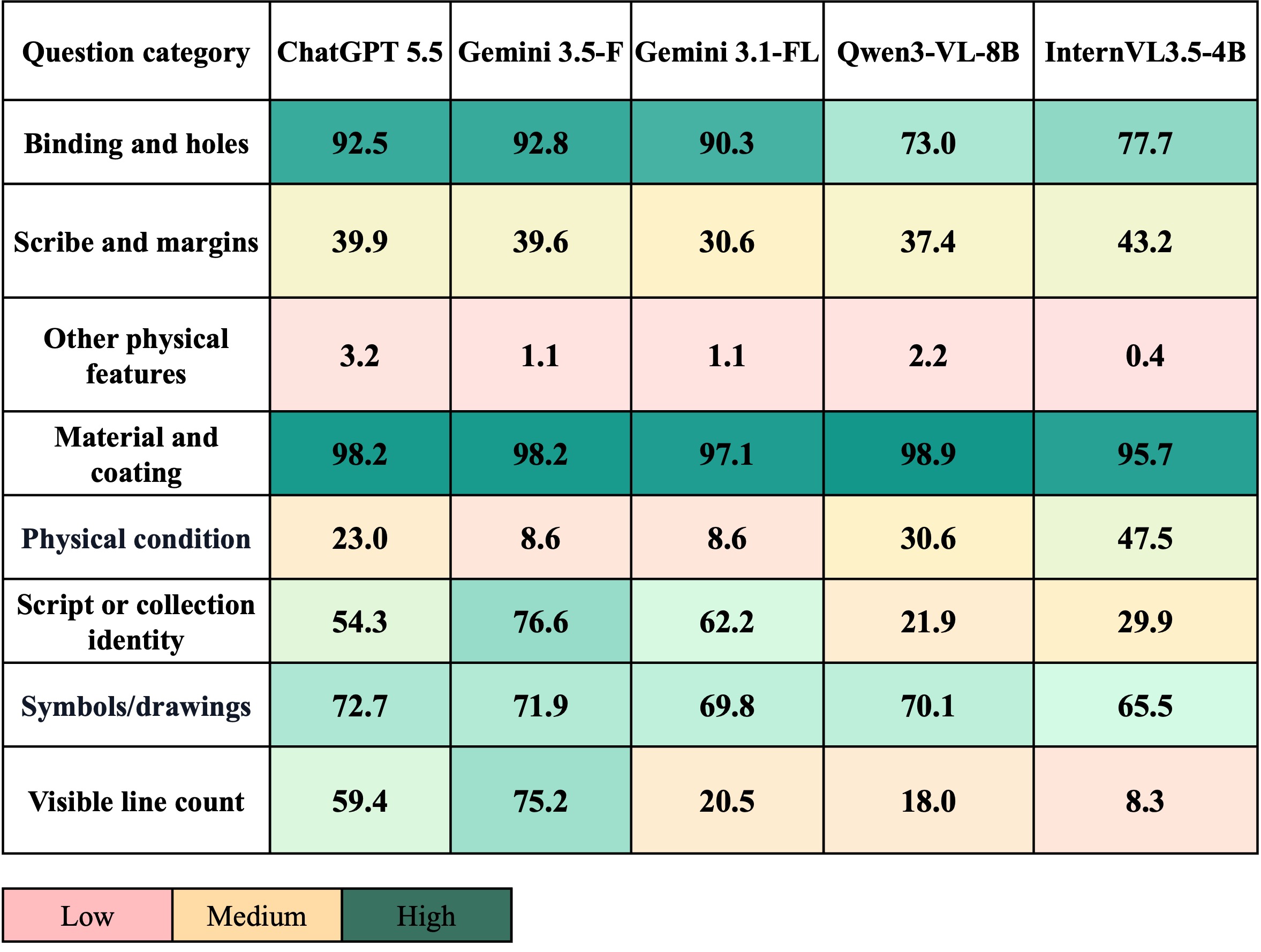}
\caption{\textbf{Question-category performance under constrained-answer prompting.} Exact-match accuracy (\%) is reported on the held-out PalmLeaf-VQA test split across the eight question categories. Rows correspond to question categories, and columns correspond to five selected high-performing models. Color intensity indicates accuracy, showing stronger performance on material and coating and binding and holes, while localized physical-feature and physical-condition reasoning remain challenging.}
\label{fig:task_family_heatmap}
\end{figure}

Figure~\ref{fig:task_family_heatmap} shows a clear split between globally visible manuscript properties and localized, interpretation-heavy cues. Across the leading constrained-answer models, categories such as material and coating and binding and holes are recognized more reliably than physical condition, margins, and other physical features. The latter categories often depend on small cracks, stains, faded regions, edge damage, or subtle non-textual marks that are easy to miss in full-page reasoning.
Line counting exhibits a different failure profile. The strongest models achieve high within-one accuracy even when exact line-count accuracy is lower, indicating that they often recover the approximate page structure but miss the reference count by one line. These errors are common when text lines are faint, broken, overlapping, or partially occluded.

Script or collection identification also remains difficult. Models must distinguish visually similar and underrepresented writing traditions that are unlikely to be well covered in generic multimodal pretraining. The collection-level results further show that this capability is highly model-dependent rather than uniformly improving with overall benchmark strength.

\subsection{Human Baseline}

We evaluate human performance on a balanced subset of \textbf{256 test questions}, with four questions sampled from each collection--category combination. Eight domain-experienced participants answer under the same constrained-answer protocol used for model evaluation, with three independent responses per question.

Human EM and F1 are averaged over all \textbf{768 responses}, while W1 is computed over the \textbf{96 responses} associated with the 32 visible-line-count questions. As shown in Table~\ref{tab:human_baseline}, humans achieve \textbf{87.50\% EM}, \textbf{92.40\% F1}, and \textbf{96.88\% W1}, with a mean pairwise exact agreement of \textbf{86.0\%}.

On the same subset, Gemini 3.5 Flash achieves \textbf{58.59\% EM}, \textbf{67.07\% F1}, and \textbf{93.75\% W1}, while ChatGPT 5.5 achieves \textbf{59.77\% EM}, \textbf{69.89\% F1}, and \textbf{90.63\% W1}. The corresponding human--model EM gaps are \textbf{28.91} and \textbf{27.73 percentage points}.

\subsection{Error Analysis}
\label{sec:error_analysis}

We observe four recurring failure modes. The first is \textbf{rare-script and collection confusion}, where visually unfamiliar scripts are mapped to more common or visually similar traditions. The second is \textbf{localized physical-feature failure}, where models overlook small cracks, stains, marginal marks, or other non-textual artifacts that are critical for preservation-oriented questions.

The third failure mode is \textbf{layout and counting ambiguity}. In many line-count questions, models predict the approximate structure correctly but fail to determine the exact number of visible lines, especially when strokes are faint or overlapping. The high W1 scores relative to exact count accuracy make this pattern particularly clear.

The fourth is \textbf{answer incompleteness and output-space drift}, which is most visible in open-answer and multi-answer settings. Models often return descriptive phrases instead of canonical labels, predict only the most salient attribute, or add plausible but unsupported secondary labels. Constrained prompting reduces these errors, but it does not remove the underlying visual-reasoning limitations.

Overall, the error profile suggests that progress on PalmLeaf-VQA will require more than stronger OCR. Effective models must combine rare-script representation, localized visual attention, preservation-aware reasoning, precise structural counting, and controlled multi-label generation.

\section{Discussion and Limitations}
\label{sec:discussion_limitations}

\subsection{Main Findings}

PalmLeaf-VQA extends document VQA beyond transcription and conventional layout analysis by evaluating historical manuscripts as both textual documents and physical artifacts. The strongest model reaches only \textbf{58.00\%} exact-match accuracy, indicating a substantial gap between general-purpose multimodal reasoning and manuscript understanding.

Constrained-answer prompting improves EM by \textbf{17.52} points on average, showing that open-answer evaluation is affected by lexical variation and output-format drift. However, the remaining gap is still large, which indicates that the main difficulty is visual rather than purely linguistic. Models continue to struggle with rare-script identification, localized damage, line counting, and multi-attribute prediction. Performance also varies substantially across collections, suggesting that script appearance, image quality, page geometry, and preservation condition all strongly affect model behavior.

\subsection{Scope and Limitations}

PalmLeaf-VQA is smaller than web-scale VQA datasets because digitized palm-leaf resources are limited, fragmented, and require substantial manual curation. The eight collection groups cover multiple traditions but do not represent the full geographic, linguistic, chronological, or material diversity of palm-leaf manuscripts. The source collections also differ in framing, resolution, background, acquisition conditions, and preservation quality. Although the benchmark is image-disjoint and related images are grouped whenever reliable source information is available, incomplete metadata prevents us from claiming fully manuscript-disjoint splits. 

In addition, the annotations focus on visually observable properties and do not cover transcription, translation, dating, authorship, or deeper philological interpretation. Exact-match accuracy is strict and reproducible, but it may penalize semantically reasonable answers that differ from the canonical representation. We therefore also report F1, Jaccard similarity, and within-one line-count accuracy. Still, these metrics do not measure explanation quality, confidence calibration, or whether the model relied on the correct image region.

\subsection{Future Directions}

Future work may extend PalmLeaf-VQA with newly digitized traditions, stronger manuscript-level provenance, richer preservation labels, region-grounded annotations, and broader expert evaluations across additional manuscript traditions. Connecting visual inspection with transcription and historical metadata would further broaden its use in cultural-heritage analysis.

\section{Conclusion}
\label{sec:conclusion}

We introduced \textbf{PalmLeaf-VQA}, a multi-script benchmark for visual question answering over historical palm-leaf manuscripts. The dataset contains \textbf{923 curated images} and \textbf{7,384 newly annotated question-answer pairs} across eight collection groups and eight manuscript-aware categories, covering classification, counting, layout reasoning, and preservation-related visual understanding.

Our evaluation of proprietary MLLMs, open-weight vision-language models, and text-only controls shows that current systems remain limited in this domain. Gemini 3.5 Flash achieves the strongest result, but reaches only \textbf{58.00\% exact-match accuracy} under constrained-answer prompting. Persistent errors in rare-script recognition, localized physical-feature analysis, line counting, and multi-attribute prediction show that the challenge extends beyond OCR and answer formatting.

PalmLeaf-VQA provides a reproducible benchmark for developing multimodal models with stronger rare-script representations, localized visual attention, structural precision, and preservation-aware reasoning for historical document analysis.

\bibliographystyle{plainnat}
\bibliography{main}

\clearpage
\section*{Supplementary Material}
\setcounter{section}{0}
\renewcommand{\thesection}{\Alph{section}}
\renewcommand{\theHsection}{supp.\Alph{section}}
\renewcommand{\theHsubsection}{supp.\Alph{section}.\arabic{subsection}}
\renewcommand{\theHsubsubsection}{supp.\Alph{section}.\arabic{subsection}.\arabic{subsubsection}}

\subsection*{Summary}

This supplementary material provides additional information that could not be included in the main paper because of space limitations. Following established practices in document VQA and multimodal benchmark design~\cite{mathew2021docvqa,fu2024ocrbenchv2improvedbenchmark,zhu2024mmdocbench,yue2026seavision}, we provide further details on dataset harmonization, annotation and validation, machine-readable data formats, prompting and model configurations, answer normalization, human evaluation, extended experiments, error analysis, licensing, and release organization.

\section{Additional Dataset Construction Details}
\label{sec:supp_dataset}

The main paper reports the collection-level composition and train, validation, and test statistics of PalmLeaf-VQA. This section therefore focuses on the curation decisions, annotation procedure, duplicate control, split-assignment strategy, machine-readable format, and release organization that are not fully described in the main paper.

\subsection{Source Harmonization}
\label{sec:supp_source_harmonization}

PalmLeaf-VQA combines manuscript images from resources originally developed for recognition, segmentation, layout analysis, enhancement, and transcription. These resources differ in image resolution, aspect ratio, framing, background, acquisition conditions, degradation level, and metadata availability.

Before annotation, all source collections were converted into a unified internal structure. Each image was assigned a benchmark identifier, original source identifier, collection group, source path, redistribution status, and available manuscript-level metadata. Original filenames and source identifiers were retained in a separate provenance file so that every benchmark item can be traced to its original source.

Images were not geometrically normalized into a fixed manuscript shape. Cropping or framing adjustment was applied only when necessary to remove unrelated borders or acquisition artifacts. Manuscript content, binding holes, margins, writing placement, degradation, and other physical characteristics were preserved because these properties are directly evaluated in PalmLeaf-VQA.

\subsection{Image Filtering and Quality Control}
\label{sec:supp_filtering}

Images were manually inspected before question annotation. An image was excluded when one or more of the following conditions applied:

\begin{itemize}
\item the manuscript was too small or blurred for reliable visual inspection;
\item a substantial part of the manuscript was outside the captured frame;
\item unrelated pages or objects dominated the manuscript content;
\item severe compression, exposure, or acquisition artifacts prevented reliable annotation;
\item the image did not provide sufficient visible evidence for the required question categories;
\item the image was an exact duplicate or a near-duplicate of another selected item.
\end{itemize}

Natural degradation was not treated as an exclusion criterion by itself. Images containing faded writing, cracks, stains, broken edges, uneven illumination, scratches, or partially damaged surfaces were retained whenever the relevant visual evidence remained interpretable. These characteristics form an important part of the benchmark difficulty.

\subsection{Duplicate and Near-Duplicate Control}
\label{sec:supp_duplicate_control}

Exact duplicates were detected through file-level comparison and image hashing. Near-duplicate candidates were identified using perceptual similarity, source filenames, collection metadata, and manual inspection.

Near-duplicate review considered whether two images represented:

\begin{itemize}
\item the same digital file stored under different names;
\item repeated captures of the same manuscript folio;
\item cropped, resized, or compressed versions of the same image;
\item consecutive captures with nearly identical visible content;
\item different folios from the same physical manuscript.
\end{itemize}

Exact duplicates were removed. Verified near-duplicates and repeated captures were grouped before split assignment to reduce visual leakage. Different folios from the same manuscript were retained when they contained meaningfully different content, but they were assigned to the same split whenever reliable manuscript-level information was available.

\subsection{Split Assignment and Leakage Prevention}
\label{sec:supp_split_assignment}

Split construction was performed after duplicate and near-duplicate screening. When reliable manuscript identifiers were available, all images associated with the same physical manuscript or capture sequence were assigned to one split. When complete identifiers were unavailable, grouping decisions used source folders, filename patterns, source metadata, perceptual similarity, and manually verified near-duplicate clusters.

The split procedure prioritized the following constraints:

\begin{enumerate}
\item no exact image appears in more than one split;
\item verified near-duplicate images remain within the same split;
\item images from the same known manuscript remain within the same split;
\item collection-level proportions remain reasonably consistent across splits;
\item every image contributes exactly one question from each of the eight question categories.
\end{enumerate}

Because complete manuscript identifiers are unavailable for some source collections, PalmLeaf-VQA is described as image-disjoint rather than fully manuscript-disjoint. The available grouping evidence for each source will be documented in the release metadata.

\subsection{Annotation Workflow and Validation}
\label{sec:supp_annotation_workflow}

The annotation team comprised 12 researchers from South, Southeast, and East Asia with experience in palm-leaf manuscripts, historical document analysis, or the represented manuscript traditions. Assignments were matched to regional, script, or collection familiarity whenever possible. This assignment strategy follows culturally grounded VQA practices that emphasize relevant linguistic and cultural expertise~\cite{romero2024cvqa}.

Each question--answer pair was written by one annotator and independently reviewed by a second team member. The reviewer checked whether the question was visually answerable, assigned to the correct category, and supported by the image evidence. Disagreements were resolved through discussion and adjudication by an experienced reviewer. Cases that remained ambiguous or visually unsupported were excluded.

The annotation workflow consisted of the following stages:

\begin{enumerate}
\item assignment of an image to an annotator with relevant collection familiarity;
\item creation of one question and answer for each benchmark category;
\item independent review by a second team member;
\item correction of category, wording, counting, or answer-format issues;
\item adjudication of disagreements;
\item final normalization of labels, counts, punctuation, and multi-label formatting.
\end{enumerate}

\subsection{Question Construction Principles}
\label{sec:supp_question_construction}

Each retained image is associated with eight questions, one from each benchmark category. Questions were designed to evaluate visually observable manuscript properties rather than external historical or linguistic knowledge.

Question construction followed four principles:

\begin{itemize}
\item \textbf{Visual grounding:} the answer must be supported directly by visible image evidence.
\item \textbf{Category consistency:} each question must evaluate its assigned benchmark category.
\item \textbf{Canonical phrasing:} semantically equivalent questions follow consistent wording patterns.
\item \textbf{Answer standardization:} categorical, binary, numerical, and multi-label answers use predefined formats.
\end{itemize}

Questions requiring full-text transcription, translation, manuscript dating, author identification, or deeper philological interpretation were excluded from the current benchmark. This design follows diagnostic benchmark practices that decompose aggregate performance into interpretable document and visual capabilities~\cite{li2024naturalbench,ouyang2025omnidocbench,zhu2024mmdocbench}.

\subsection{Answer Types and Canonicalization}
\label{sec:supp_answer_design}

PalmLeaf-VQA contains single-label, multi-label, binary, and numerical answers. The answer format is determined by the question category and visible evidence.

\begin{table*}[t]
\centering
\small
\setlength{\tabcolsep}{5pt}
\begin{tabular}{@{}lll@{}}
\toprule
\textbf{Question category} &
\textbf{Primary answer format} &
\textbf{Output rule} \\
\midrule
Script Name & Single label & One canonical collection-group label \\
Physical Condition & Single label & One canonical condition label \\
Line Count & Count & One non-negative integer \\
Symbols \& Drawings & Single label & Standardized \texttt{yes}/\texttt{no} answer \\
Material \& Coating & Single label & One canonical material or coating label \\
Binding \& Holes & Count & One non-negative integer for visible binding holes \\
Scribe \& Margins & Multi-label & One or more canonical layout labels \\
Other Physical Features & Multi-label & One or more canonical physical-feature labels \\
\bottomrule
\end{tabular}
\caption{\textbf{Answer formats used in PalmLeaf-VQA.}
Formats match the released \texttt{annotation/answer\_vocabularies.json} file.}
\label{tab:supp_answer_formats}
\end{table*}

Multi-label answers are stored as unordered canonical sets. Binary answers use standardized \texttt{yes} and \texttt{no} values. Visible-line-count answers are stored as integers. Synonymous or variant labels are mapped to canonical representations during normalization.

\subsection{Annotation Record Structure}
\label{sec:supp_json}

Each question--answer pair is stored as an independent JSON record linked to its manuscript image. The record contains benchmark identifiers, question metadata, the reference answer, and the split assignment. Provenance, licensing, redistributability, checksums, and constrained-answer vocabularies are retained in separate artifact files.

\begin{table*}[t]
\centering
\small
\setlength{\tabcolsep}{5pt}
\begin{tabular}{@{}lll@{}}
\toprule
\textbf{Field} & \textbf{Type} & \textbf{Description} \\
\midrule
\texttt{sample\_id}      & String         & Unique question--answer identifier used by the benchmark \\
\texttt{image\_name}     & String         & Image filename in the corresponding collection folder \\
\texttt{script\_name}    & String         & One of the eight PalmLeaf-VQA collection groups \\
\texttt{question\_id}    & String/Integer & Question index or identifier for the image \\
\texttt{task\_family}    & String         & Higher-level task family for analysis \\
\texttt{category}        & String         & Manuscript-aware question category \\
\texttt{question\_type}  & String         & Classification, count, binary, layout, or attribute-recognition type \\
\texttt{question\_en}    & String         & English natural-language question \\
\texttt{answer}          & String         & Canonical reference answer after annotation normalization \\
\texttt{split}           & String         & Training, validation, or test split \\
\bottomrule
\end{tabular}
\caption{\textbf{PalmLeaf-VQA QA annotation fields.}
The table follows the released \texttt{annotation/qa\_schema.json}. Source
provenance, license status, redistributability, and checksums are retained in
separate metadata and manifest files.}
\label{tab:supp_json_fields}
\end{table*}

An illustrative annotation record is shown below:

{\scriptsize
\begin{verbatim}
{
  "sample_id": "Balinese_0002__1",
  "image_name": "Balinese_0002.jpg",
  "script_name": "Balinese",
  "question_id": 1,
  "task_family": "Script recognition",
  "category": "Script Name",
  "question_type": "classification",
  "question_en": "What is the script name?",
  "answer": "Balinese",
  "split": "test"
}
\end{verbatim}
}

For multi-label questions, answers are stored as unordered canonical label sets. Binary answers use standardized \texttt{yes} and \texttt{no} values, while visible-line-count answers are stored as integers. The illustrative record follows the released QA schema; complete records are provided in the train, validation, test, and sample files.

\subsection{Source Provenance and Release Organization}
\label{sec:supp_release_structure}

The release package separates benchmark annotations from source-specific image permissions. For redistributable resources, curated images and annotations will be released together. For restricted resources, the release will contain question--answer annotations, source identifiers, retrieval instructions, and processing scripts.

The provenance file records:

\begin{itemize}
\item benchmark and original source identifiers;
\item source collection and citation;
\item original access location;
\item redistribution status;
\item license or usage condition;
\item split assignment;
\item available manuscript or capture-sequence identifier;
\item integrity checksum for the corresponding image.
\end{itemize}

This organization allows the benchmark annotations and evaluation tools to remain reproducible without redistributing images whose original access conditions do not permit direct redistribution.

\subsection{Representative Supplementary Sample}
\label{sec:supp_sample}

A representative sample of 80 manuscript images and 640 question--answer records is included in the supplementary archive. The sample covers all eight collection groups, all eight question categories, the major answer formats, and varied manuscript conditions.

The sample demonstrates the dataset organization and annotation quality. It does not replace the complete benchmark release. Images are included only when redistribution is permitted; otherwise, the sample contains the corresponding annotations, source identifiers, and retrieval information.

\section{Experimental Setup and Prompting}
\label{sec:supp_setup}

The use of open-answer and constrained-answer conditions follows recent multimodal evaluation protocols that distinguish visual-understanding errors from lexical variation, instruction following, and output-format errors~\cite{liu2024mmbench,yue2024mmmupro,fu2024ocrbenchv2improvedbenchmark}.

\subsection{Hardware Environment}
\label{sec:supp_hardware}

Open-weight models were evaluated on a server equipped with two NVIDIA A800 GPUs. Each GPU provides 80GB of memory, giving 160GB of total available GPU memory. Depending on the model and memory requirements, inference used one GPU or a multi-GPU device mapping.

Proprietary models were evaluated through their corresponding APIs and therefore did not use the local A800 hardware for model execution.

\begin{table}[t]
\centering
\small
\setlength{\tabcolsep}{5pt}
\begin{tabular}{@{}p{0.36\linewidth}p{0.56\linewidth}@{}}
\toprule
\textbf{Component} & \textbf{Configuration} \\
\midrule
Local GPUs & $2\times$ NVIDIA A800 80GB PCIe \\
Aggregate GPU memory & 160GB \\
Driver / CUDA report & NVIDIA driver 580.126.09; CUDA 13.0 \\
Local precision & bfloat16 when supported \\
Batch size & 1 image-question group per forward pass \\
API execution & Proprietary models evaluated remotely through APIs \\
Artifact environment & Python 3.11; \texttt{pytest}; \texttt{pyyaml} \\
\bottomrule
\end{tabular}
\caption{\textbf{Hardware and software environment used for evaluation.}
Open-weight inference used local A800 GPUs. Proprietary API models did not use
local GPU memory for model execution.}
\label{tab:supp_hardware}
\end{table}

\subsection{Evaluation Conditions}
\label{sec:supp_conditions}

Model--prompt runs were evaluated on the same 2,224 test questions under open-answer and constrained-answer prompting. The manuscript image, question, and reference annotation remained unchanged between the two conditions.

In the open-answer condition, the model generated a free-form response without receiving category-specific answer choices. In the constrained-answer condition, the model additionally received the required output format and, where applicable, the canonical answer vocabulary.

All experiments were conducted zero-shot. Models did not receive task-specific fine-tuning, in-context demonstrations, OCR transcripts, source metadata, reference labels, or preservation annotations. The two text-only controls received the question but not the manuscript image.

\subsection{Model and Inference Configurations}
\label{sec:supp_model_configs}

Table~\ref{tab:supp_model_configs} summarizes the evaluated model families and decoding defaults recorded in the release artifact.

\begin{table*}[t]
\centering
\scriptsize
\setlength{\tabcolsep}{4pt}
\begin{tabular}{@{}lllll@{}}
\toprule
\textbf{Model} & \textbf{Access} & \textbf{Prompt modes} &
\textbf{Default decoding} & \textbf{Execution} \\
\midrule
Gemini 3.5 Flash & Proprietary API & Open, constrained & temperature 0; top-$p$ 1; max 1024 tokens & Remote API \\
ChatGPT 5.5 & Proprietary API & Open, constrained & JSON schema; max 1200 tokens & Remote API \\
Gemini 3.1 Flash-Lite & Proprietary API & Open, constrained & temperature 0; top-$p$ 1; max 256 tokens & Remote API \\
Qwen3-VL-8B-Instruct & Open-weight & Open, constrained & deterministic; max 256 tokens & Local A800 GPUs \\
Qwen2.5-VL-7B-Instruct & Open-weight & Open, constrained & deterministic; max 256 tokens & Local A800 GPUs \\
Qwen2.5-VL-3B-Instruct & Open-weight & Open, constrained & deterministic; max 256 tokens & Local A800 GPUs \\
InternVL3.5-8B & Open-weight & Open, constrained & deterministic; max 256 tokens & Local A800 GPUs \\
InternVL3.5-4B & Open-weight & Open, constrained & deterministic; max 256 tokens & Local A800 GPUs \\
Gemma-3-4B-IT & Open-weight & Open, constrained & deterministic; max 256 tokens & Local A800 GPUs \\
Llama-3.1-8B-Instruct & Text-only control & Open, constrained & deterministic; max 256 tokens & Local A800 GPUs \\
Llama-3.2-3B-Instruct & Text-only control & Open, constrained & deterministic; max 256 tokens & Local A800 GPUs \\
\bottomrule
\end{tabular}
\caption{\textbf{Model and inference configuration summary.}
The evaluated model list and decoding defaults match
\texttt{configs/models.yaml}. The two Llama models are text-only controls and
do not receive manuscript images.}
\label{tab:supp_model_configs}
\end{table*}

For each model, the released configuration additionally records:

\begin{itemize}
\item API access date or checkpoint revision;
\item library and SDK version;
\item temperature, top-$p$, and maximum output tokens;
\item batch size and device mapping;
\item image resizing or tiling configuration;
\item random seed, where applicable;
\item retry policy for API or inference failures;
\item treatment of empty, refused, or invalid responses.
\end{itemize}

\subsection{Open-Answer Prompt}
\label{sec:supp_open_prompt}

The open-answer condition provides the image and question without
category-specific answer choices. The Gemini 3.5 Flash open-answer notebook used
the following prompt template, with \texttt{\{question\}} replaced by the natural
language question:

\begin{quote}
\small\ttfamily
Answer the question based only on the image.\\[2pt]
Rules:\\
- Use only visible information in the image.\\
- Do not use external knowledge.\\
- Do not transcribe the manuscript text unless the question explicitly asks for transcription.\\
- Do not translate the manuscript text.\\
- Give the shortest direct answer supported by the image.\\
- Do not explain your answer.\\
- Do not add extra text before or after the JSON.\\
- Return exactly one valid JSON object.\\
- For a single answer, use: \{\textquotedbl{}answer\textquotedbl{}: \textquotedbl{}short answer\textquotedbl{}\}.\\
- If the question asks for multiple visible items or features, use: \{\textquotedbl{}answer\textquotedbl{}: [\textquotedbl{}item1\textquotedbl{}, \textquotedbl{}item2\textquotedbl{}]\}.\\
- If the answer is unknown or not visible, use: \{\textquotedbl{}answer\textquotedbl{}: \textquotedbl{}unknown\textquotedbl{}\}.\\[2pt]
Question: \{question\}\\
\end{quote}

The exact notebook-derived prompt is released as
\texttt{prompts/gemini\_open\_answer.txt}. A shorter backend-agnostic description
of the same open-answer condition is also provided in
\texttt{prompts/open\_answer\_template.txt}.

\subsection{Constrained-Answer Prompt}
\label{sec:supp_constrained_prompt}

The constrained-answer condition uses the same manuscript image and question
content, but additionally provides category-specific output rules and allowed
answer vocabularies. For Gemini, each question is answered as one JSON object
with a single \texttt{answer} key. The notebook-derived prompt template is
released in \texttt{prompts/gemini\_constrained\_answer.txt}; the concrete prompt
appends the appropriate allowed-answer block for the question category.

For the OpenAI/ChatGPT constrained run, one manuscript image is sent with all
questions for that image. The OpenAI runner used:

\begin{quote}
\scriptsize\ttfamily
Answer the questions about the palm-leaf manuscript image.\\
Return JSON only with this shape:\\
\{\textquotedbl{}answers\textquotedbl{}:[\{\textquotedbl{}question\_id\textquotedbl{}:\textquotedbl{}...\textquotedbl{},\\
\textquotedbl{}answer\textquotedbl{}:\textquotedbl{}...\textquotedbl{}\}]\}.\\
Keep answers short and direct. Use the image and question text.\\
If uncertain, make the best visual estimate.\\[2pt]
Questions:\\
\{questions\_json\}
\end{quote}

The OpenAI API run additionally requested a JSON schema requiring an
\texttt{answers} array, where each item contains string-valued
\texttt{question\_id} and \texttt{answer} fields. The exact template and schema
are released in \texttt{prompts/constrained\_answer.txt}.

\subsection{Category-Specific Output Rules}
\label{sec:supp_output_rules}

The constrained-answer instructions were adapted to the corresponding answer type:

\begin{itemize}
\item \textbf{Single-label questions:} return one canonical label.
\item \textbf{Multi-label questions:} return all applicable labels as a comma-separated unordered set.
\item \textbf{Binary questions:} return only \texttt{yes} or \texttt{no}.
\item \textbf{Visible-line-count questions:} return one integer without additional text.
\end{itemize}

The image and question remain identical between open-answer and constrained-answer evaluation. Only the answer instruction and available vocabulary differ. The term \textit{Closed} is used only as a compact label in selected figures.

\subsection{Prediction Record Format}
\label{sec:supp_prediction_format}

Each model output was stored together with its model identifier, prompt condition, raw response, normalized response, reference answer, and metric values.

{\scriptsize
\begin{verbatim}
{
  "sample_id": "Balinese_0002__1",
  "model_name": "Qwen3-VL-8B-Instruct",
  "prompt_condition": "constrained",
  "raw_prediction": "Balinese",
  "normalized_prediction": "balinese",
  "ground_truth": "Balinese",
  "exact_match": 1,
  "precision": 1.0,
  "recall": 1.0,
  "f1": 1.0,
  "jaccard": 1.0
}
\end{verbatim}
}

API failures, empty outputs, refusals, and unparseable responses were logged rather than silently removed.

\section{Answer Normalization and Evaluation Metrics}
\label{sec:supp_evaluation}

Consistent with document VQA and OCR-oriented benchmarks~\cite{mathew2021docvqa,liu2024ocrbench,fu2024ocrbenchv2improvedbenchmark,zhu2024mmdocbench}, predictions are normalized before exact-match and partial-credit evaluation.

\subsection{Normalization Procedure}
\label{sec:supp_normalization}

Predictions and references were normalized before scoring. The procedure includes:

\begin{itemize}
\item conversion to lowercase;
\item removal of leading and trailing whitespace;
\item repeated-whitespace normalization;
\item punctuation cleanup where punctuation is not semantically relevant;
\item canonicalization of script and collection names;
\item standardization of binary labels;
\item numerical parsing for visible-line-count questions;
\item conversion of multi-label responses into unordered canonical sets;
\item removal of repeated labels;
\item rejection of invalid or unparseable outputs.
\end{itemize}

Normalization does not add labels or change the semantic content of an answer. Predictions outside the allowed vocabulary in the constrained-answer setting are treated as incorrect unless they map unambiguously to a documented canonical alias.

\begin{table*}[t]
\centering
\small
\setlength{\tabcolsep}{5pt}
\begin{tabular}{@{}lll@{}}
\toprule
\textbf{Raw output} & \textbf{Normalized output} & \textbf{Rule} \\
\midrule
\texttt{BALINESE.} & \texttt{balinese} & Case and punctuation normalization \\
\texttt{yes, it is visible} & \texttt{yes} & Binary-answer canonicalization \\
\texttt{twelve} & \texttt{12} & Documented numerical mapping \\
\texttt{label\_b, label\_a} & \texttt{\{label\_a,label\_b\}} & Unordered multi-label set \\
\texttt{label\_a, label\_a} & \texttt{\{label\_a\}} & Duplicate-label removal \\
\bottomrule
\end{tabular}
\caption{\textbf{Illustrative normalization examples.}
The released \texttt{annotation/normalization\_rules.yaml} and
\texttt{eval/normalize.py} define the exact normalization behavior used for
scoring.}
\label{tab:supp_normalization_examples}
\end{table*}

\subsection{Exact-Match Accuracy}
\label{sec:supp_em}

Exact-match accuracy is defined as:

\begin{equation}
\mathrm{EM}=\frac{1}{N}\sum_{i=1}^{N}\mathbb{1}\left[\operatorname{norm}(\hat{a}_i)=\operatorname{norm}(a_i)\right].
\end{equation}

where $\hat{a}_i$ is the predicted answer, $a_i$ is the reference answer, and $\operatorname{norm}(\cdot)$ denotes answer normalization.

For multi-label questions, exact match requires equality between the complete predicted and reference sets.

\subsection{Partial-Credit Metrics}
\label{sec:supp_partial_metrics}

For a predicted label set $P$ and reference set $G$, precision, recall, F1, and Jaccard similarity are defined as:

\begin{equation}
\mathrm{Precision}=\frac{|P\cap G|}{|P|},\qquad
\mathrm{Recall}=\frac{|P\cap G|}{|G|}.
\end{equation}

\begin{equation}
\mathrm{F1}=\frac{2\cdot\mathrm{Precision}\cdot\mathrm{Recall}}{\mathrm{Precision}+\mathrm{Recall}}.
\end{equation}

and

\begin{equation}
\mathrm{Jaccard}=\frac{|P\cap G|}{|P\cup G|}.
\end{equation}

The released scoring code explicitly defines the behavior for empty predicted or reference sets.

\subsection{Line-Count Metrics}
\label{sec:supp_count_metrics}

Count-exact accuracy requires the predicted integer to equal the reference count. Within-one accuracy is defined as:

\begin{equation}
\mathrm{W1}=\frac{1}{N_c}\sum_{i=1}^{N_c}\mathbb{1}\left[|\hat{c}_i-c_i|\leq 1\right].
\end{equation}

where $N_c$ is the number of visible-line-count questions, $\hat{c}_i$ is the predicted count, and $c_i$ is the reference count.

\section{Human Baseline Protocol}
\label{sec:supp_human}

\subsection{Balanced Subset Construction}
\label{sec:supp_human_subset}

Human evaluation was conducted on a balanced subset of 256 test questions. The subset contains four questions from each collection--category combination:

\begin{equation}
8\ \text{collection groups}\times 8\ \text{question categories}\times 4\ \text{questions}=256.
\end{equation}

This design gives equal representation to all collection groups and question categories and avoids overrepresenting larger collections or easier question types. The stratified design follows the fine-grained evaluation principles used in recent multimodal benchmarks~\cite{li2024naturalbench,ouyang2025omnidocbench}.

\subsection{Participants and Assignment}
\label{sec:supp_human_participants}

Eight domain-experienced participants completed the evaluation. Participants had experience in palm-leaf manuscripts, historical document analysis, or the represented manuscript traditions.

Each question received three independent responses. Participants did not evaluate examples for which they produced the final reference annotation. Anonymous participant identifiers were used in the stored response files.

\subsection{Evaluation Instructions}
\label{sec:supp_human_instructions}

Participants received the manuscript image, question, and the same constrained-answer vocabulary supplied to the evaluated models. They were instructed to use only visible image evidence and not to rely on external historical or linguistic knowledge.

Single-label questions required one canonical answer. Binary questions required \texttt{yes} or \texttt{no}. Numerical questions required one integer. Multi-label questions permitted all applicable canonical labels.

\subsection{Human-Score Aggregation}
\label{sec:supp_human_aggregation}

Three responses were collected for each of the 256 questions, producing 768 individual response-level predictions. Human EM and F1 were calculated by scoring each response independently against the reference and averaging across all 768 responses. Majority voting was not used.

The balanced subset contains 32 visible-line-count questions. Human W1 was therefore computed over:

\begin{equation}
32\ \text{questions}\times 3\ \text{responses}=96\ \text{line-count responses}.
\end{equation}

Model EM and F1 were computed once over the 256-question subset, while model W1 was computed once over the corresponding 32 visible-line-count questions.

\subsection{Response Agreement}
\label{sec:supp_human_agreement}

Response consistency was measured using mean pairwise exact agreement after applying the same normalization rules used for model evaluation.

Let $y_{i,r}$ denote the normalized response from participant $r$ for question $i$. With three responses per question, the pairwise agreement score is:

\begin{equation}
A_i =
\frac{1}{3}
\sum_{(r,s)\in\{(1,2),(1,3),(2,3)\}}
\mathbb{1}
\left[
\operatorname{norm}(y_{i,r})=\operatorname{norm}(y_{i,s})
\right].
\end{equation}

Category-level agreement is obtained by averaging $A_i$ over questions within each category. Overall agreement is computed as the macro-average across the eight categories. This formulation is appropriate because PalmLeaf-VQA contains heterogeneous categorical, binary, numerical, and multi-label answer spaces.

\subsection{Human Baseline Results}
\label{sec:supp_human_results}

\begin{table}[t]
\centering
\scriptsize
\setlength{\tabcolsep}{3.5pt}
\begin{tabular}{@{}lrrrrr@{}}
\toprule
\textbf{Evaluator} &
\textbf{Questions} &
\textbf{Responses} &
\textbf{EM} &
\textbf{F1} &
\textbf{W1} \\
\midrule
Human participants & 256 & 768 & 87.50 & 92.40 & 96.88 \\
Gemini 3.5 Flash & 256 & 256 & 58.59 & 67.07 & 93.75 \\
ChatGPT 5.5 & 256 & 256 & 59.77 & 69.89 & 90.63 \\
\bottomrule
\end{tabular}
\caption{\textbf{Human and model performance on the balanced comparison subset.}
Human EM and F1 are averaged over 768 individual responses, while human W1 is
computed over 96 responses to the 32 visible-line-count questions. Model EM and
F1 are evaluated over 256 questions, and model W1 is evaluated over the
corresponding 32 visible-line-count questions. Values are percentages.}
\label{tab:supp_human_results}
\end{table}

Human participants achieve 87.50\% EM, 92.40\% F1, and 96.88\% W1, with a mean pairwise exact agreement of 86.0\%. The corresponding human--model EM gaps are 28.91 percentage points for Gemini 3.5 Flash and 27.73 percentage points for ChatGPT 5.5.

\subsection{Human Evaluation Release}
\label{sec:supp_human_release}

The public release will contain, where permitted:

\begin{itemize}
\item identifiers for the 256 sampled questions;
\item anonymous participant identifiers;
\item raw and normalized responses;
\item per-response EM, F1, and W1 values;
\item category-level agreement statistics;
\item participant assignment information;
\item human-evaluation instructions.
\end{itemize}

No participant names or personally identifying information will be released.

\section{Extended Experimental Results}
\label{sec:supp_extended_results}

The main paper reports the principal benchmark results. The supplementary material provides complementary analyses without repeating the same summary tables.

\subsection{Complete Category-Level Results}
\label{sec:supp_category_results}

The supplementary results report all evaluated models across the eight question categories under both prompting conditions. These tables include exact-match accuracy and the corresponding task-specific complementary metrics where applicable.

The machine-readable complete by-category tables are released in \texttt{results/per\_category\_metrics.csv}.

\subsection{Complete Collection-Level Results}
\label{sec:supp_collection_results}

The main paper reports constrained-answer collection-level accuracy. The supplement additionally reports open-answer collection-level results and, where informative, F1 or Jaccard similarity for each collection group.

The machine-readable collection-level tables are released in \texttt{results/per\_category\_metrics.csv}.

\subsection{Answer-Cardinality Analysis}
\label{sec:supp_cardinality_results}

Results are reported separately for single-answer and multi-answer questions. Multi-answer questions include exact match, precision, recall, F1, and Jaccard similarity because a model may recover only a subset of the required labels.

\subsection{Line-Count Error Analysis}
\label{sec:supp_line_count_results}

The supplement reports count-exact accuracy, W1, and the distribution of absolute counting errors. A histogram of $|\hat{c}-c|$ distinguishes near-miss predictions from substantial layout failures.

\subsection{Additional Ablations}
\label{sec:supp_ablations}

Where supported by completed experiments, the following ablations are included:

\begin{itemize}
\item open-answer versus constrained-answer prompting;
\item format instructions without a vocabulary versus instructions with a vocabulary;
\item scoring before and after answer normalization;
\item alternative image-resolution settings for selected open-weight models;
\item image-conditioned models versus text-only controls.
\end{itemize}

The completed open-versus-constrained and text-only-control comparisons are reported in the machine-readable result tables.

\section{Extended Error Analysis}
\label{sec:supp_error_analysis}

This section combines aggregate and qualitative analyses of model failures, following the diagnostic evaluation style of recent document and multimodal benchmarks~\cite{fu2024ocrbenchv2improvedbenchmark,zhu2024mmdocbench,yue2026seavision}. The aggregate analysis identifies the categories in which incorrect predictions occur most frequently, while the qualitative analysis illustrates how prompt format, visual ambiguity, localized evidence, and answer cardinality affect individual predictions.

\subsection{Aggregate Error Distribution}
\label{sec:supp_error_distribution}

\begin{figure}[t]
\centering
\includegraphics[width=\linewidth]{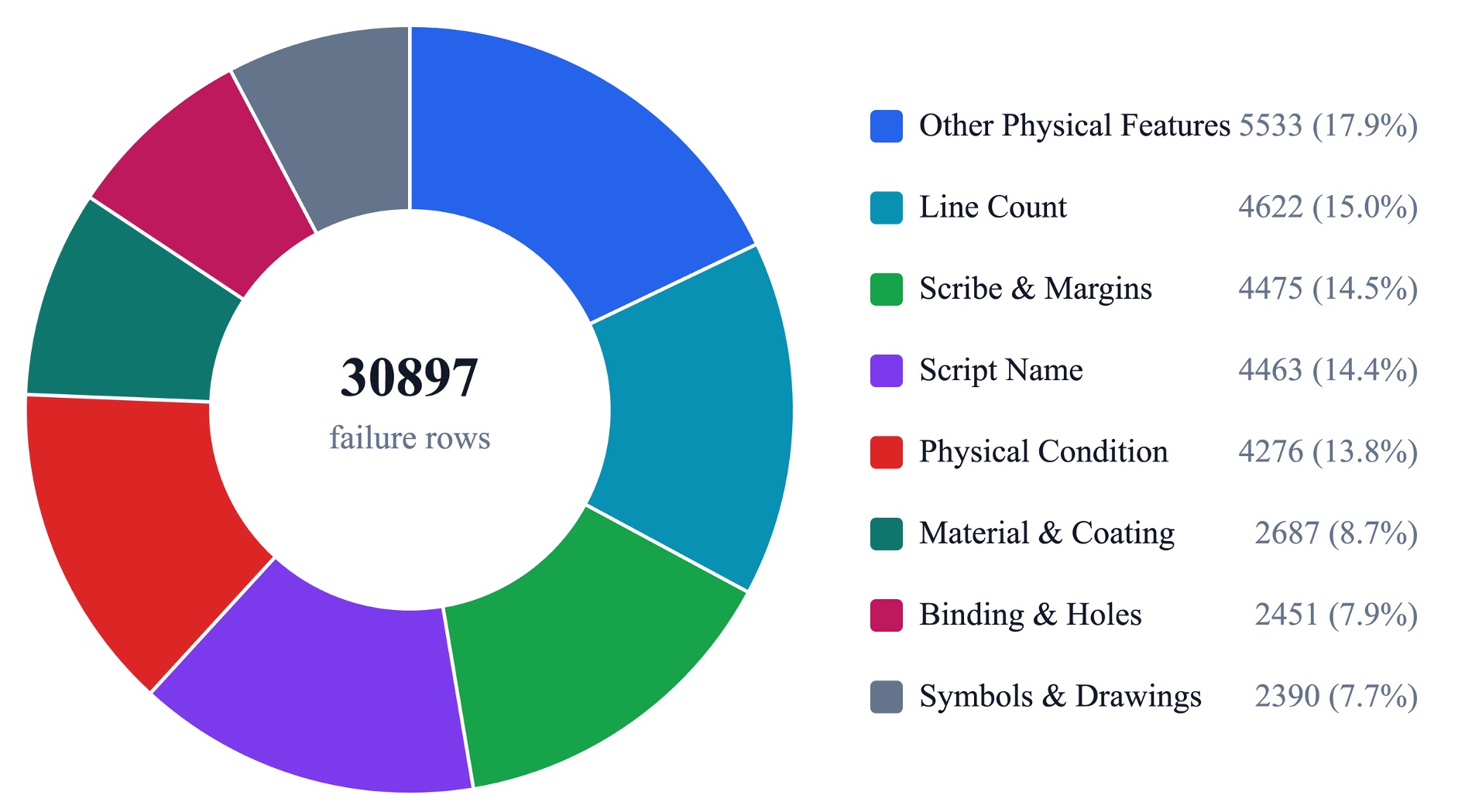}
\caption{\textbf{Aggregate distribution of incorrect predictions across PalmLeaf-VQA question categories.}
The chart pools 30,897 incorrect row-level predictions from 20 available model--prompt runs. Counts and percentages indicate each category's contribution to the total number of exact-match failures. Summary-only runs without row-level prediction files are excluded. Visible-line-count failures are determined using exact-count correctness rather than W1.}
\label{fig:supp_error_distribution}
\end{figure}

Figure~\ref{fig:supp_error_distribution} summarizes incorrect predictions across the eight PalmLeaf-VQA question categories. Because each evaluated model receives the same number of questions from each category, category-level failure counts are directly comparable within a fixed set of models and prompting conditions.

The aggregate figure is generated directly from the final raw prediction files, and its caption states the scope of the included model--prompt runs, prediction rows, and exact-match failures.

The released row-level error distribution is computed over the available prediction-level CSV files. It includes 20 model--prompt runs, 44,480 prediction rows, and 30,897 exact-match errors.

The error distribution indicates whether failures are concentrated in categories requiring localized feature recognition, precise structural counting, or fine-grained collection discrimination. Other physical features frequently require models to recover several low-contrast attributes, while visible-line-count questions are sensitive to faint, fragmented, overlapping, or partially occluded lines.

\begin{figure*}[t]
\centering
\includegraphics[width=\textwidth]{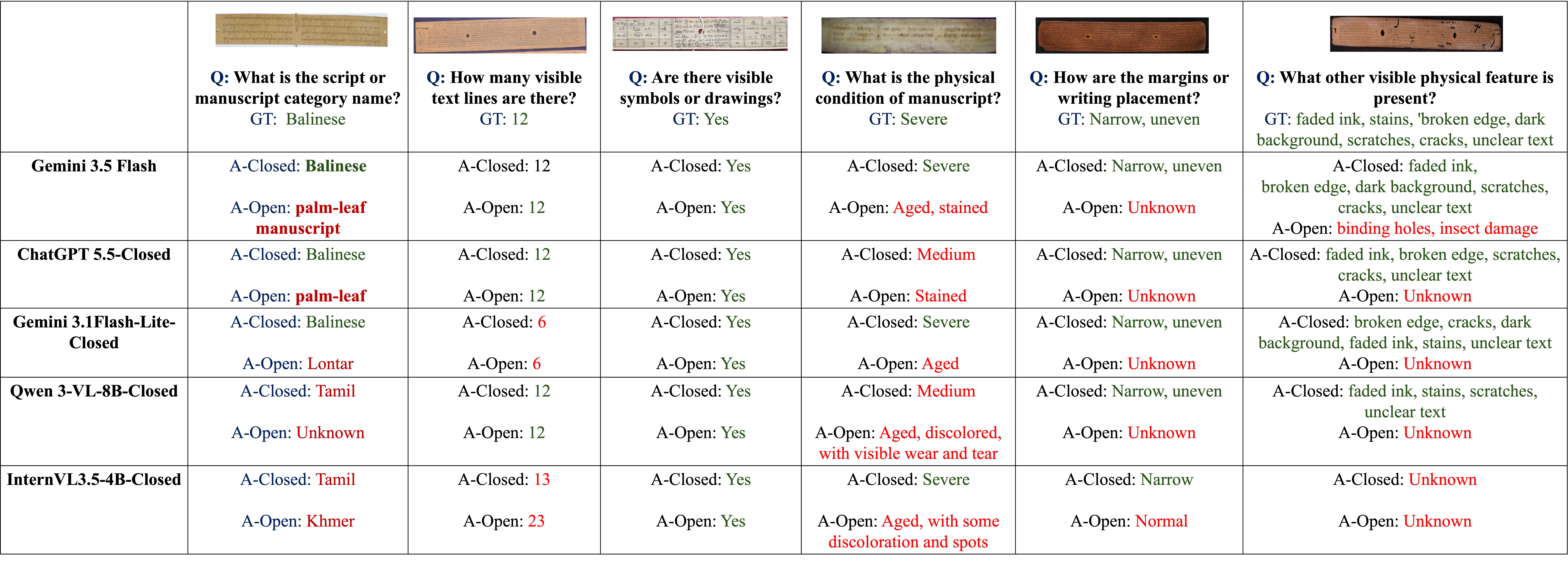}
\caption{\textbf{Qualitative comparison of open-answer and constrained-answer predictions.}
Each card includes the manuscript image, question category, question, reference answer, model prediction, and exact-match status. Green denotes reference-consistent outputs, whereas red denotes incorrect, incomplete, unsupported, or non-canonical responses. The examples illustrate correct canonical predictions, line-count errors, localized physical-feature failures, and open-answer output drift.}
\label{fig:supp_open_closed_examples}
\end{figure*}
\subsection{Qualitative Comparison of Prompting Conditions}
\label{sec:supp_prompt_qualitative}

Figure~\ref{fig:supp_open_closed_examples} shows representative correct and incorrect predictions from the representative sample subset. Each card contains a manuscript image, question, ground-truth answer, model prediction, and exact-match status.

The examples show that constrained-answer prompting reduces lexical variation and output-space drift. Open-answer responses may use broad or non-canonical descriptions such as ``palm-leaf manuscript,'' ``aged,'' or ``unknown'' instead of the required benchmark labels. Constrained prompting improves answer consistency but does not remove visual-reasoning errors.

The examples illustrate four recurring patterns. First, models confuse visually similar or underrepresented manuscript traditions. Second, localized attributes such as cracks, stains, faded regions, marginal marks, and broken edges are often missed. Third, models may recover the approximate page structure while predicting an incorrect line count. Fourth, multi-label responses may omit part of the reference set or add unsupported attributes.

In the visualization, green text denotes reference-consistent predictions. Red text denotes incorrect, incomplete, unsupported, or non-canonical outputs.

\subsection{Summary of Failure Patterns}
\label{sec:supp_error_summary}

The combined quantitative and qualitative analyses identify four dominant failure modes:

\begin{itemize}
\item \textbf{Rare-script and collection confusion:} models map unfamiliar manuscript traditions to more common or visually similar groups.
\item \textbf{Localized physical-feature failure:} small cracks, stains, scratches, faded regions, and marginal details are overlooked.
\item \textbf{Layout and counting ambiguity:} models recover approximate structure but fail to determine the exact number of visible lines.
\item \textbf{Answer incompleteness and output drift:} open-answer and multi-label responses omit required attributes, introduce unsupported labels, or use non-canonical wording.
\end{itemize}

These findings show that progress on PalmLeaf-VQA requires more than stronger OCR. Effective models must combine rare-script representation, localized visual attention, preservation-aware reasoning, precise structural counting, and controlled multi-label generation.
\section{Licensing, Ethics, and Release}
\label{sec:supp_licensing}

\subsection{Intended and Unsupported Uses}
\label{sec:supp_uses}

PalmLeaf-VQA is intended for research on multimodal document understanding, historical manuscript analysis, visual question answering, rare-script representation, preservation-aware reasoning, and benchmark evaluation.

The benchmark is not intended for historical authentication, ownership decisions, automated conservation intervention, cultural attribution without expert review, or complete transcription and translation.

\subsection{Cultural-Heritage Considerations}
\label{sec:supp_cultural}

Palm-leaf manuscripts may contain culturally, historically, or religiously significant material. Benchmark predictions should not replace manuscript experts, conservators, source institutions, or relevant communities. Users should respect the access conditions and cultural context of the original collections.

\subsection{Planned Release Components}
\label{sec:supp_release_components}
Upon publication of the paper, we plan to release the PalmLeaf-VQA dataset and associated resources, subject to source-specific redistribution permissions. The release will include benchmark results, evaluation code, prompts, and documentation through the public GitHub repository \url{https://github.com/back-kh/PalmLeaf-VQA} and Hugging Face.

\end{document}